\documentclass[10pt,twocolumn,letterpaper]{article}

\usepackage{wacv}  
\usepackage[accsupp]{axessibility}

\definecolor{wacvblue}{rgb}{0.21,0.49,0.74}
\usepackage[pagebackref,breaklinks,colorlinks,allcolors=wacvblue]{hyperref}
\usepackage{wrapfig}
\usepackage{makecell}
\usepackage{booktabs}
\usepackage{siunitx}
\usepackage[table]{xcolor} 
\usepackage{amssymb}
\def\wacvPaperID{1957} 
\def\confName{WACV}
\def\confYear{2026}

\title{DMS2F-HAD: A Dual-branch Mamba-based Spatial–Spectral Fusion Network for Hyperspectral Anomaly Detection}

\author{Aayushma Pant\qquad Lakpa Tamang\qquad Tsz-Kwan Lee\qquad Sunil Aryal\\
School of Information Technology, Deakin University\\
Waurn Ponds, Geelong, Victoria, Australia\\
{\tt\small \{a.pant, l.tamang, glory.lee, sunil.aryal\}@deakin.edu.au}
}

\begin{document}
\maketitle
\begin{abstract}

Hyperspectral anomaly detection (HAD) aims to identify rare and irregular targets in high-dimensional hyperspectral images (HSIs), which are often noisy and unlabeled data. Existing deep learning methods either fail to capture long-range spectral dependencies (e.g., convolutional neural networks) or suffer from high computational cost (e.g., Transformers). To address these challenges, we propose DMS2F-HAD, a novel dual-branch Mamba-based model. Our architecture utilizes Mamba’s linear-time modeling to efficiently learn distinct spatial and spectral features in specialized branches,which are then integrated by a dynamic gated fusion mechanism to enhance anomaly localization. Across fourteen benchmark HSI datasets, our proposed DMS2F-HAD not only achieves a state-of-the-art average AUC of 98.78\%, but also demonstrates superior efficiency with an inference speed $4.6\times$ faster than comparable deep learning methods. The results highlight DMS2F-HAD's strong generalization and scalability, positioning it as a strong candidate for practical HAD applications. The source code, models  are available at:\\
\noindent \textcolor{blue}{\url{https://github.com/Ayushma00/DMS2F-HAD}}
\end{abstract}
           
\section{Introduction}
\label{sec:intro}

\begin{figure}[t]
  \centering
   \includegraphics[width=0.8\columnwidth]{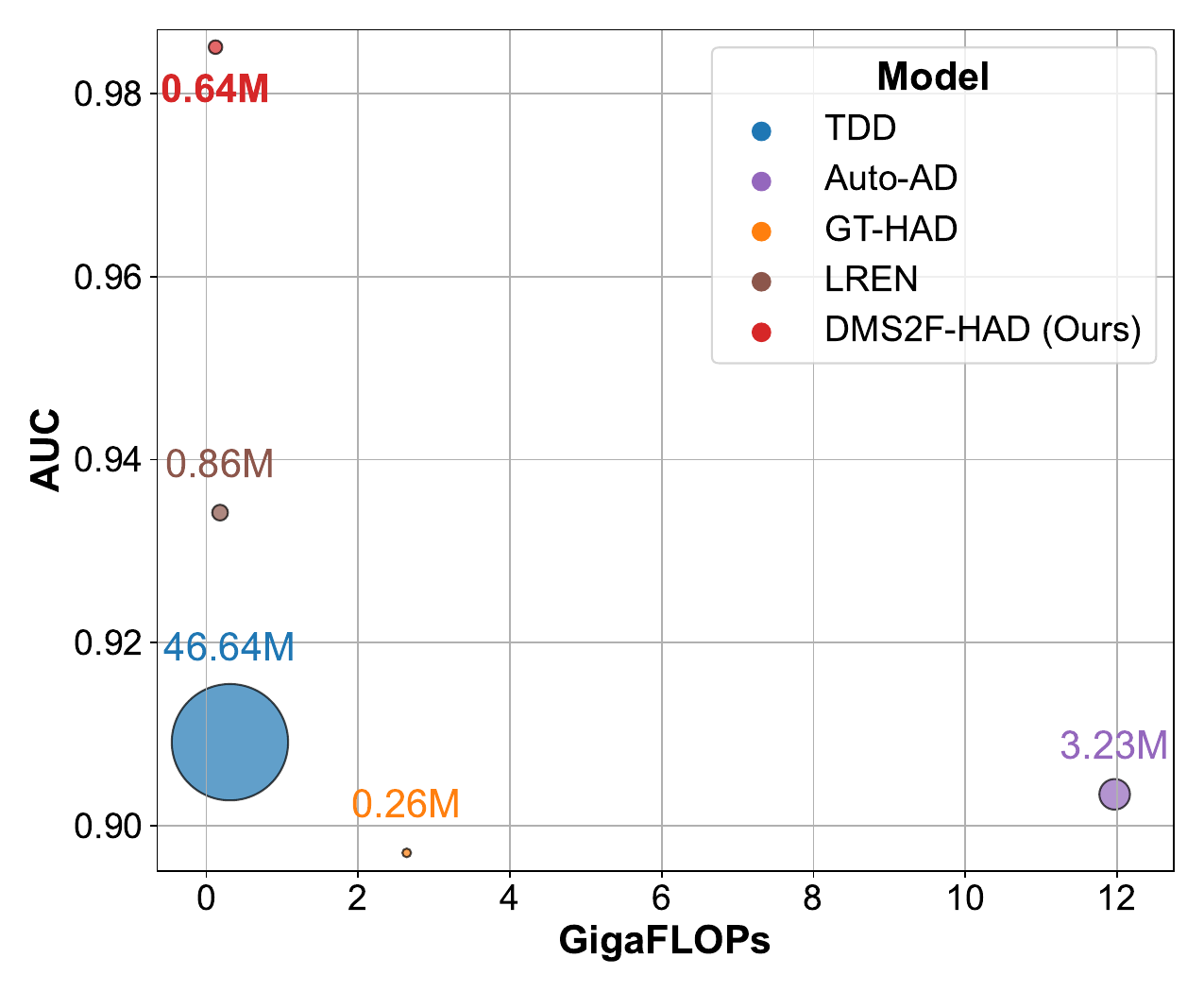}
   \caption{The accuracy-efficiency trade-off on the AVIRIS-2 dataset, where circle size reflects model parameters. Our DMS2F-HAD clearly achieves the best balance, delivering the highest AUC with minimal complexity.
}
\label{fig:model_complexity}
\end{figure}
Hyperspectral images (HSIs) are high-dimensional datasets that capture rich spatial and spectral information across a wide range of the electromagnetic spectrum \cite{7564440}. This makes them effective for identifying anomalous targets, which are rare and irregular objects that exhibit significant spectral deviations from the background and have a lower probability of occurrence. These anomalies often include man-made objects such as aircraft, space debris, rooftops, and vehicles, varying according to the surrounding context. Hyperspectral anomaly detection (HAD) is an area of research to detect such anomalies. HAD is utilised as an unsupervised target detection approach across various applications, including remote sensing, military surveillance, mineral exploration, and search and rescue operations.
\par Traditionally, a significant body of research was carried out using statistical approaches \cite{9521674, 60107, 6851148, 1386510}. Typically, such methods operate under a naive assumption that the background data follows a certain statistical distribution (e.g. multivariate Gaussian). Although they can be fast, one of the major downsides is that they often yield high false positives in complex scenes, i.e. the background is full of many classes like vehicles, roof, tree, grass, water, road, path, and shadow, and when assumptions are violated \cite{9930793}. Another approach to realise HAD is through representation-based methods \cite{9826842,9444588,7875485, Ling2019ACS,9288702} where a dictionary of background signatures is learned to separate anomalies from the background. Although these methods offer improved outlier separability \cite{9532003}, they are often sensitive to noise, require high computational resources, and involve manual parameter tuning. Furthermore, their detection performance heavily depends on parameter selection, making them less suitable for scalable and practical deployment. 
\par 

While Deep Learning (DL) methods like CNNs \cite{https://doi.org/10.1155/2015/258619,9423171, 10049187} and Transformers \cite{10432978,10073635,10506667,dosovitskiy2021imageworth16x16words} have improved HAD, they face significant trade-offs. CNNs struggle with long-range spectral dependencies due to limited receptive fields, while Transformers suffer from quadratic computational complexity, making them unsuitable for resource-constrained real-time applications.
Furthermore, the extensive number of parameters inherent in their architecture makes it prone to overfitting, particularly when the training data is imbalanced. These shortcomings limit the application of HAD in the real world \cite{wang2025acmambafastunsupervisedanomaly}. In addition to algorithmic limitations, effective fusion of spatial and spectral information remains a key challenge in HAD. Although spatial features capture local structures, and spectral features provide material signatures. Many methods aim to integrate both modalities but overemphasise the spectral domain and neglect the spatial correlations \cite{9040873,9521674,10858750}, leading to poor anomaly localization and higher false positives.
\par To address these gaps, we propose DMS2F-HAD, a novel dual-branch Mamba-based network tailored for hyperspectral anomaly detection. While inspired by the efficient spectral grouping strategy used in classification tasks \cite{10812905}, our work effectively adapts this mechanism for unsupervised background reconstruction. The main contributions of this paper are summarized as follows:

\begin{itemize}
    \item \textbf{Unsupervised Dual-Branch Mamba Autoencoder:} We propose the first dual-branch Mamba-based autoencoder specifically designed for the reconstruction-based anomaly detection task. Unlike classification models \cite{10812905} that fuse features for discrimination, we introduce a lightweight \textit{Spatial-Spectral Decoder} that uses Mamba's linear complexity to accurately reconstruct background clutter while suppressing anomalous targets.

    \item \textbf{Adaptive Gated Fusion Mechanism:} We design a content-aware gated fusion mechanism that dynamically arbitrates between spatial and spectral branches. Distinct from static summation or the channel-wise modulation used in prior works, our learnable gate ($G$) explicitly weighs spatial texture against spectral consistency pixel-by-pixel, enhancing the model's ability to minimize false alarms in complex, textured backgrounds.

    \item \textbf{Superior Accuracy-Efficiency Trade-off:} Validated on fourteen benchmark datasets, DMS2F-HAD achieves a state-of-the-art average AUC of 98.78\%. Crucially,as shown in Figure~\ref{fig:model_complexity}, it demonstrates remarkable efficiency, operating 4.6$\times$ faster than Transformer-based methods and requiring 3.3$\times$ fewer parameters than the leading Mamba-based anomaly detector (MMR-HAD)\cite{10884568}, making it highly suitable for resource-constrained onboard processing.
\end{itemize}
\section{Background and Related Works}
\subsection{Hyperspectral Anomaly Detection}
\label{sec:had}
Research on HAD has evolved from traditional statistical methods to modern DL techniques including CNNs \cite{7875485, 9423171, 10049187,pant2025hyperspectralanomalydetectionmethods}, autoencoders (AEs) \cite{9382262}, and generative adversarial networks (GANs) \cite{10.1016/j.neunet.2024.107036}. While CNNs often demand large labeled datasets and overlook spectral dependencies, autoencoders are sensitive to noise and complex backgrounds, and GANs face high training costs. Recent advancements in Transformer-based methodologies, such as GT-HAD \cite{10432978} and STAD \cite{MA2025107036}, have incorporated self-attention mechanisms and anomaly-aware feature learning to effectively capture subtle spectral variations. However, there reliance on attention mechanism leads to a quadratic increase in computational complexity with sequence length, posing a significant bottleneck for high-dimensional HSI data. Alternative approaches, including transferred direct detection (TDD) \cite{10506667} and teacher–student frameworks \cite{9157778,zhang2023destsegsegmentationguideddenoising}, aim to bypass reconstruction processes or distil knowledge from a teacher network. Nevertheless, they still require computationally intensive architectures. Despite these advancements, most existing methods struggle to jointly capture sequential spectral correlations and spatial contextual cues inherent to HSIs, under the constraints of real-time inference and generalisability across datasets. These challenges underscore the necessity for architectures that are both computationally efficient and sequence-aware. Mamba, a recent state-space model, provides linear complexity and rapid inference for long-range sequential modeling. We extend this model through a dual-branch design with adaptive gated fusion, specifically tailored for HAD.

\subsection{State Space Models}
State space models (SSMs) \cite{gu2020hipporecurrentmemoryoptimal,NEURIPS2021_05546b0e,gu2022efficientlymodelinglongsequences} have recently demonstrated strong performance in capturing long-range dependencies in vision tasks. Unlike transformers, which suffer from quadratic complexity due to the attention mechanism, SSMs utilise linear-time recurrent architectures based on state-space representations, making them well-suited for long-range data such as HSIs, which can have hundreds of spectral bands. While traditional SSMs were formulated as recurrent systems, which limited training parallelism, recent advancements have unlocked their potential for highly efficient, hardware-aware architectures.



\subsection{Mamba Models}
Mamba \cite{gu2024mambalineartimesequencemodeling,ruan2024vmunetvisionmambaunet,zhu2024visionmambaefficientvisual,hu2024zigmaditstylezigzagmamba} is a recent advancement in the SSM family that uses a selective scan mechanism to selectively focus on relevant information in a sequence while operating with linear-time complexity. This design makes it uniquely suited for processing the long sequences inherent in HSI data, as it is both computationally efficient and highly parallelizable on modern hardware. 
\par While Mamba was initially a 1D sequence model, architectures like Vision Mamba \cite{zhu2024visionmambaefficientvisual}, VMamba \cite{liu2024vmambavisualstatespace} have successfully adapted it for 2D spatial data in general vision tasks\cite{zhang2024surveyvisualmamba, zhu2024visionmambaefficientvisual}. Its application to HSI analysis is nascent but promising, with early works demonstrating success in classification \cite{10740056, yao2024spectralmambaefficientmambahyperspectral}. However, its potential in HAD remains underexplored.


While MMR-HAD  \cite{10884568} recently introduced Mamba to HAD, it relies on a hybrid architecture incorporating computationally intensive multi-scale attention mechanisms. In contrast, our work proposes a purely SSM-based dual-branch autoencoder. We uniquely decouple spatial and spectral modeling into specialized, lightweight Mamba branches fused by a learnable gate, thereby eliminating the quadratic bottlenecks of attention layers entirely while achieving superior efficiency.

\section{The Proposed Method}
\label{sec:proposed_method}

Building upon Mamba, a state-space model characterized by linear complexity and efficient long-range sequence processing, we have developed a dual-branch architecture featuring adaptive gated fusion. This architecture is specifically designed to simultaneously capture spectral correlations and spatial contextual cues in hyperspectral data. The design ensures computational efficiency while providing strong anomaly detection capabilities in a variety of datasets. In the sections that follow, we offer a comprehensive analysis of the architecture and illustrate the successful application of anomaly detection within this framework.

\begin{figure*}[htbp]
\centerline{\includegraphics[width=0.8\linewidth]{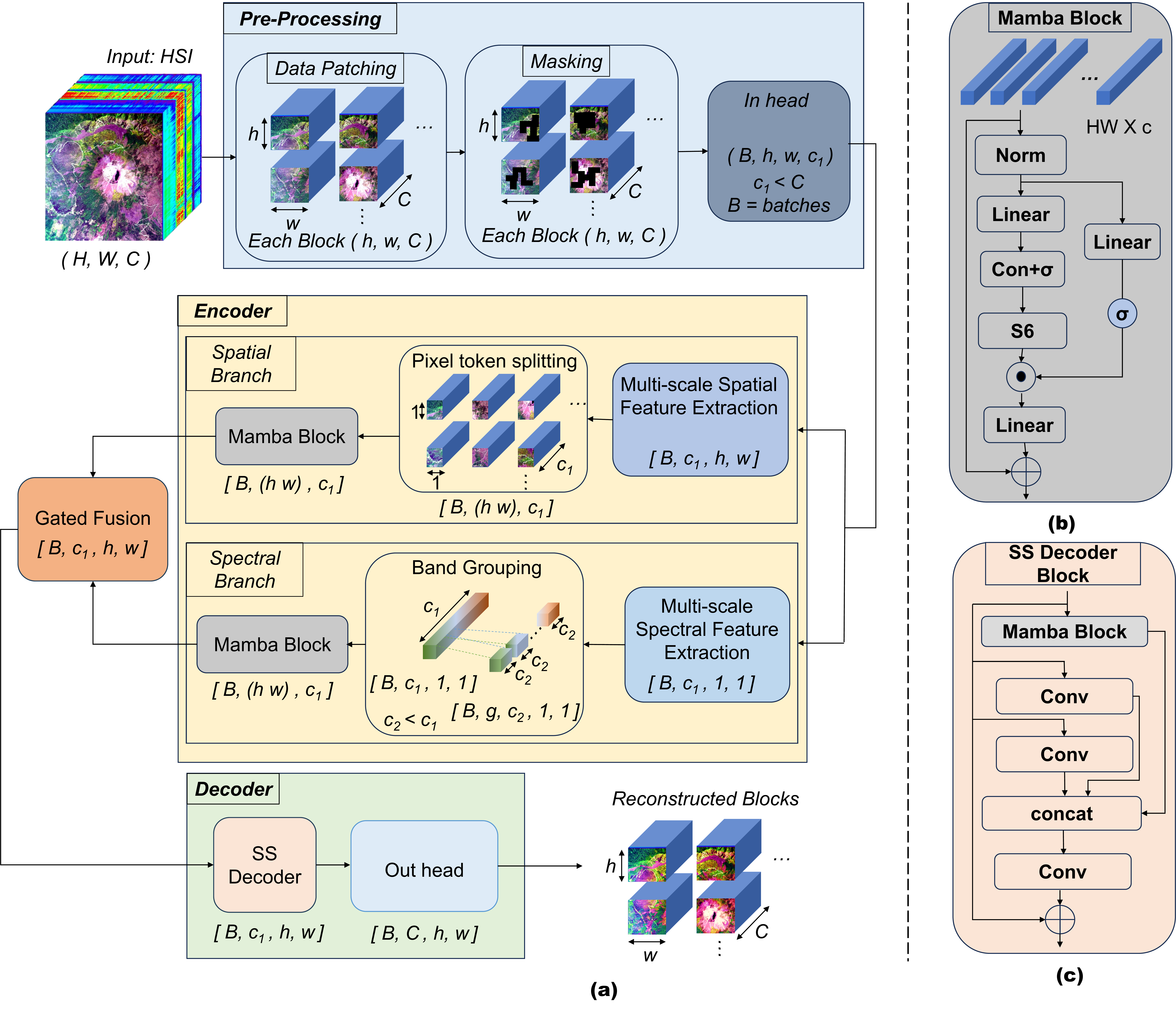}}
\vspace{-5pt}
\caption{(a) Overall architecture of  DMS2F-HAD framework; (b) Mamba block; and (c) SS Decoder Block}
\label{fig:architecture}
\end{figure*}




\subsection{Overall Framework}
The architecture of {DMS2F-HAD} is depicted in Figure ~\ref{fig:architecture}. Our architecture consists of four main components: (i) \textbf{Data processing pipeline}, (ii) \textbf{A Dual-branch Encoder} (spatial and spectral Mamba modules), (iii) \textbf{An adaptive fusion mechanism}, and (iv) \textbf{A Decoder}.
\subsubsection{Data Pre-Processing }
Given a hyperspectral image 
\(\mathbf{X}\in\mathbb{R}^{H\times W\times C}\), 
where \(H\), \(W\), and \(C\) denote its height, width, and number of spectral bands, respectively, we begin by extracting \(N\) overlapping 3D patches $\{\mathbf{x}_i\}_{i=1}^N$ using a sliding window of size $h\times w$ and stride $s$, yielding $\mathbf{x}_i\in\mathbb{R}^{h\times w\times C}$. This patching step (i) increases the number of effective training samples via overlapping crops, (ii) preserves local spatial–spectral structure critical for HAD, and (iii) reduces memory/compute so high-resolution HSIs can be processed in mini-batches.
\par Next, we employ a random spatial masking technique \cite{10005599,10884568} during training to enhance feature learning. Here, for each patch, a rectangular region is masked out across all spectral bands with a predefined probability. This forces the model to learn meaningful representations by inferring missing information from the surrounding context, thereby improving its generalization capability for handling occluded or corrupted data in real-world scenarios. Note that the size and position of the mask are randomized for each training sample to ensure diversity. Finally, the masked patches are then projected into a lower-dimensional embedding space using an input projection layer, with a \(1 \times 1\) convolution, followed by Batch Normalisation and GELU activation function. With this projection, the spectral channel dimension reduces from $C$ to a smaller dimension $c_1$, creating shallow feature maps $\mathbf{F}_i \in \mathbb{R}^{h \times w \times c_1}$. We use these feature maps as the input tokens to the dual-branch encoder which we describe in the following section.

\subsubsection{Dual-branch Encoder Block}

In the Dual-branch Encoder, we use spatial branch to capture contextual information and a spectral branch to model band-wise correlations.
\vspace{-4pt}
\paragraph{Spatial Branch:}
This branch processes the spatial context of each patch. First, the input feature map $\mathbf{F}_i$ is passed through a Multi-Scale Feature Extraction (MSFE) module composed of parallel $3 \times 3$ and $5 \times 5$ 2D convolutions, which capture spatial features at different scales. The resulting feature map $F^{\text{spa}}_{\text{ms}}$, is then flattened into $n$ spatial tokens following a row-major order. While this linearizes the 2D structure, Mamba’s Selective Scan (S6) mechanism (Figure~\ref{fig:architecture} (b)) maintains an effective global receptive field. This allows the model to capture spatial dependencies and structural anomalies across the entire image patch sequence, regardless of the linear distance between pixels in the flattened array. The output of this block is the spatial feature map $F_{\text{spa}} \in \mathbb{R}^{h \times w \times c_1}$.

\paragraph{Spectral Branch:}While the spatial branch focuses on contextual textures, the spectral branch is designed to model long-range band correlations. Each pixel in the patch contains a spectral vector of length $c_1$, which can be treated as a one-dimensional sequence. Directly feeding this long sequence to a Mamba block is computationally inefficient and does not exploit local spectral smoothness inherent in HSIs. 
To address this, we apply a Spectral Grouping strategy. We segment the spectral dimension $c_1$ into $g$ overlapping sub-sequences of length $c_2$ with stride $k$. The adjacent groups share $c_2 - k$ bands. This overlap preserves spectral continuity at group boundaries, ensuring that material signatures spanning across groups are not lost. Each group is processed as a sequence by Mamba, capturing local and global spectral dependencies with linear complexity before being projected back to the original dimension to form the final spectral feature map $F_{\text{spe}} \in \mathbb{R}^{h \times w \times c_1}$.

\subsubsection{Adaptive Gated Fusion}
After dual-branch encoding of the spatial and spectral tokens, we merge the outputs of both branches, via an adaptive gated fusion mechanism. Unlike simple concatenation, which merely stacks spatial and spectral features without considering their relative contributions, often leading to redundant representations and suboptimal feature utilization our gated fusion method is designed to dynamically adjust the weighting of spatial and spectral features for each pixel. We represent the fused representation by $F_{\text{fusion}}$ which is computed as follows:
\begin{equation}
\begin{aligned}
F_{\text{fusion}} &= \text{Proj} \left( \mathbf{G} \odot F_{\text{spa}} + (1 - \mathbf{G}) \odot F_{\text{spe}} \right)
\end{aligned}
\label{eq:4}
\end{equation}
where, $\mathbf{G} = \sigma \left( \text{Conv}_{\text{gate}} \left( [F_{\text{spa}}, F_{\text{spe}}] \right) \right)$ is the learnable gated network. Here, the channel-wise concatenated features are passed through \(\text{Conv}_{\text{gate}}\), a \(1 \times 1\) convolution layer with activation function, $\sigma$. \(\text{Proj}(\cdot)\) is a \(1 \times 1\) fusion projection layer. As shown in our ablation study \ref{tab:ablation_all}, this is critical for higher accuracy as it allows the network to prioritize spatial texture in heterogeneous urban areas while focusing on spectral consistency in homogeneous backgrounds, a flexibility that static summation lacks.



\subsection{Decoder and Reconstruction}
Next, the fused feature \(F_{\text{fusion}}\) from Eq.~\eqref{eq:4} is forwarded to a lightweight decoder to reconstruct the original HSI patch. Our decoder utilizes a Spatial-Spectral Decoder (SS Decoder) that mirrors the encoder's ability to process global and local information. The schematic representation of SS Decoder is shown in Figure~\ref{fig:architecture} (c). It is comprised of a Mamba block to capture the global context and parallel $3 \times 3$ and $5 \times 5$ convolution layers to restore fine-grained spatial details. The outputs from the Mamba and convolution paths are concatenated. Finally, a $1 \times 1$ convolutional head projects the features back to the original spectral dimension $C$, yielding the reconstructed patch $\hat{\mathbf{x}}_i \in \mathbb{R}^{h \times w \times C}$. The full reconstructed HSI, \(\hat{\mathbf{X}} \in \mathbb{R}^{H \times W \times C}\), is obtained by aggregating all reconstructed patches, averaging the values in the overlapping regions. We use this full patch of HSI to perform anomaly detection in the next section.

\subsection{Anomaly Detection}
Anomaly detection is performed based on the reconstruction error. We compute the pixel-wise residual error map $\mathbf{R} \in \mathbb{R}^{H \times W}$ by calculating the $L_2$ norm between the original and reconstructed spectral vectors:
\begin{equation}
R(i,j) = \left\| \mathbf{X}(i,j,:) - \hat{\mathbf{X}}(i,j,:) \right\|_2
\label{eq:5}
\end{equation}
This approach is based on the assumption that the model, trained on background data, will accurately reconstruct normal (background) pixels while failing to reconstruct anomalous pixels, which deviate from the learned distribution. Our DMSF2 architecture significantly enhances this process. The encoder effectively captures long-range spatial–spectral dependencies while suppressing irrelevant variations. Concurrently, the decoder employs these structured representations to more accurately reconstruct background pixels. This makes anomalous regions whose spatial or spectral correlations are inconsistent with the learned background stand out more clearly in the residual error map. Consequently, pixels with high residual errors are flagged as anomalies.

\begin{figure*}[t]
    \centering
    \includegraphics[width=0.8\linewidth]{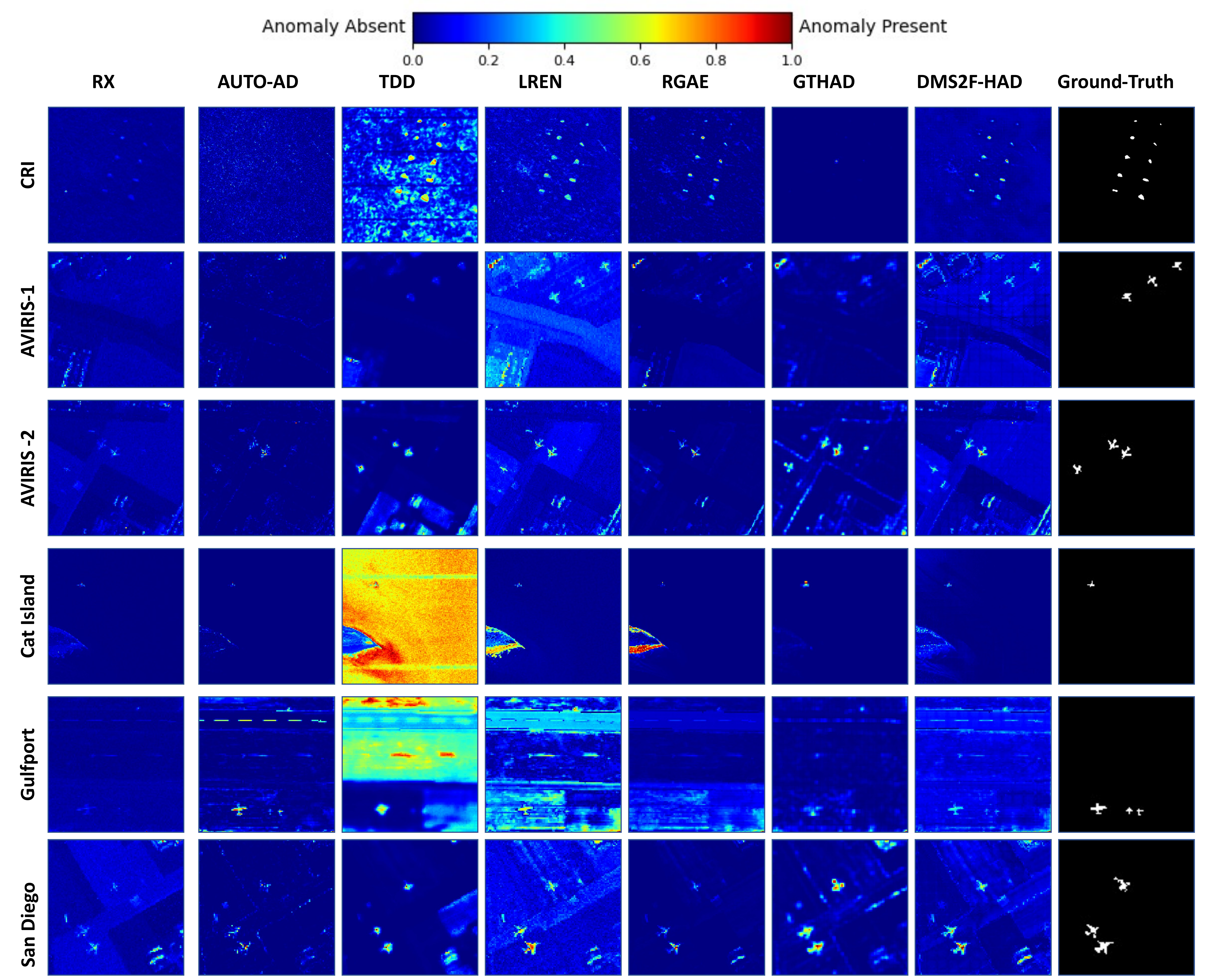}
    \caption{Colour Anomaly maps of different HAD methods on six datasets}
    \label{fig:anomaly} 
\end{figure*}

\section{Experiments and Analysis} 
\label{sec:5_experiments}

\subsection{Datasets}
We utilize fourteen benchmark HSI datasets commonly used in the literature, covering a wide range of scenes, including urban, vegetation, and coastal areas, and feature diverse anomaly types and sizes.  A summary of these datasets is provided in Table~\ref{tab:dataset_summary}. All datasets were captured with the AVIRIS sensor, except CRI, which used the Nuance CRI sensor.
\begin{table}[!ht]
\centering
\small 
\setlength{\tabcolsep}{2pt}
\begin{tabular}{l c c l}
\toprule
\textbf{Dataset} & \textbf{Size ($H \times W \times C$)} & \textbf{Res. (m)} & \textbf{Anomaly} \\
\midrule
CRI \cite{10506667} & 400$\times$400$\times$46 & --    & Rocks   \\ 
Salinas \cite{mpp2-my34-23}     & 512$\times$217$\times$224 & 3.7   & Rooftops \\
San Diego \cite{8833502}& 100$\times$100$\times$189 & 3.5   & Planes     \\
Bay Champagne \cite{7994698}  & 100$\times$100$\times$188 & 4.4   & Watercraft \\
Abu-urban-2 \cite{7994698}   & 100$\times$100$\times$207 & 17.2  & Buildings \\
Abu-urban-3 \cite{7994698}    & 100$\times$100$\times$191 & 3.5   & Buildings \\
Abu-urban-4 \cite{7994698}   & 100$\times$100$\times$205 & 7.1   & Houses \\
Abu-urban-5 \cite{7994698}   & 100$\times$100$\times$205 & 7.1   & Houses \\
AVIRIS-1 \cite{10506667}& 100$\times$100$\times$186 & 3.5   & Planes   \\
AVIRIS-2 \cite{10506667}& 128$\times$128$\times$189 & 3.5   & Plane    \\ 
Cat Island \cite{10884568}& 150$\times$150$\times$188 & 17.2  & Plane     \\
Gulfport \cite{10884568}& 100$\times$100$\times$191 & 3.4   & Planes     \\

Pavia University \cite{9684450} & 224$\times$423$\times$102 & 1.3   & Vehicles \\
Texas Coast \cite{GUO2024104200}      & 100$\times$100$\times$204 & 17.2  & Buildings \\

\bottomrule
\end{tabular}
\caption{An overview of the benchmark HAD datasets used for evaluation. ``Res." refers to spatial resolution (meters per pixel).}
\label{tab:dataset_summary}
\end{table}


\subsection{Baselines}
We benchmark DMS2F-HAD model against several state-of-the-art (SOTA) methods to ensure a comprehensive evaluation. The selected baselines include the classic statistical benchmark, RX\cite{60107}, and deep learning-based models: TDD\cite{10506667}, a transformer-based detector; GT-HAD\cite{10432978}, which uses a gated transformer; RGAE\cite{9494034}, a robust graph autoencoder; AUTO-AD\cite{9382262}, an autoencoder-based method; and LREN\cite{jiang_lren_2021}, which employs a low-rank representation. This diverse set allows us to compare against statistical, reconstruction, and transformer-based approaches. Our selection criteria prioritize methods with publicly available code to ensure fair and reproducible comparisons. While we acknowledge recent Mamba-based work like MMRHAD\cite{10884568}, the lack of a public implementation at the time of our experiments prevented its inclusion in our full benchmark evaluation. However, we provide a comparison with reported results in common datasets in Section~\ref{sec:Mambaeval}.

\subsection{Training Configuration}
All deep learning models were trained and evaluated on an NVIDIA A100 GPU. For fair comparison, we implemented all methods, including our own, under a similar computational setup. We used Adam optimizer with a learning rate of $5 \times 10^{-4}$ and a weight decay of $1 \times 10^{-4}$ for 100 epochs on each dataset. Unless otherwise stated, we set patch size $h,w=16$, embedding dimension $c_1=64$, group size channels $c_2=16$, and stride $k=8$. Empirical tests indicated the model is robust to minor variations in these parameters, provided $c_2$ is sufficient to capture local spectral continuity.We report test results based on the model with the best validation performance. 


\subsection{Evaluation}
We assess performance of all HAD methods through qualitative and quantitative approaches. In the former, we use color anomaly maps to show background suppression, and box-whisker plots to evaluate the separability between anomaly and background scores. For quantitative evaluation, we use two metrics: (i) Receiver Operating Characteristic (ROC) curve; that plots the probability of detection against the false alarm rate, and (ii) the Area Under the ROC Curve (AUC); a robust metric well-suited for class-imbalanced anomaly detection. In the following sections, we discuss in details regarding the results across these metrics.

\subsubsection{Qualitative Analysis}
\begin{figure*}[!htb]
    \centering
    \includegraphics[width=0.8\textwidth]{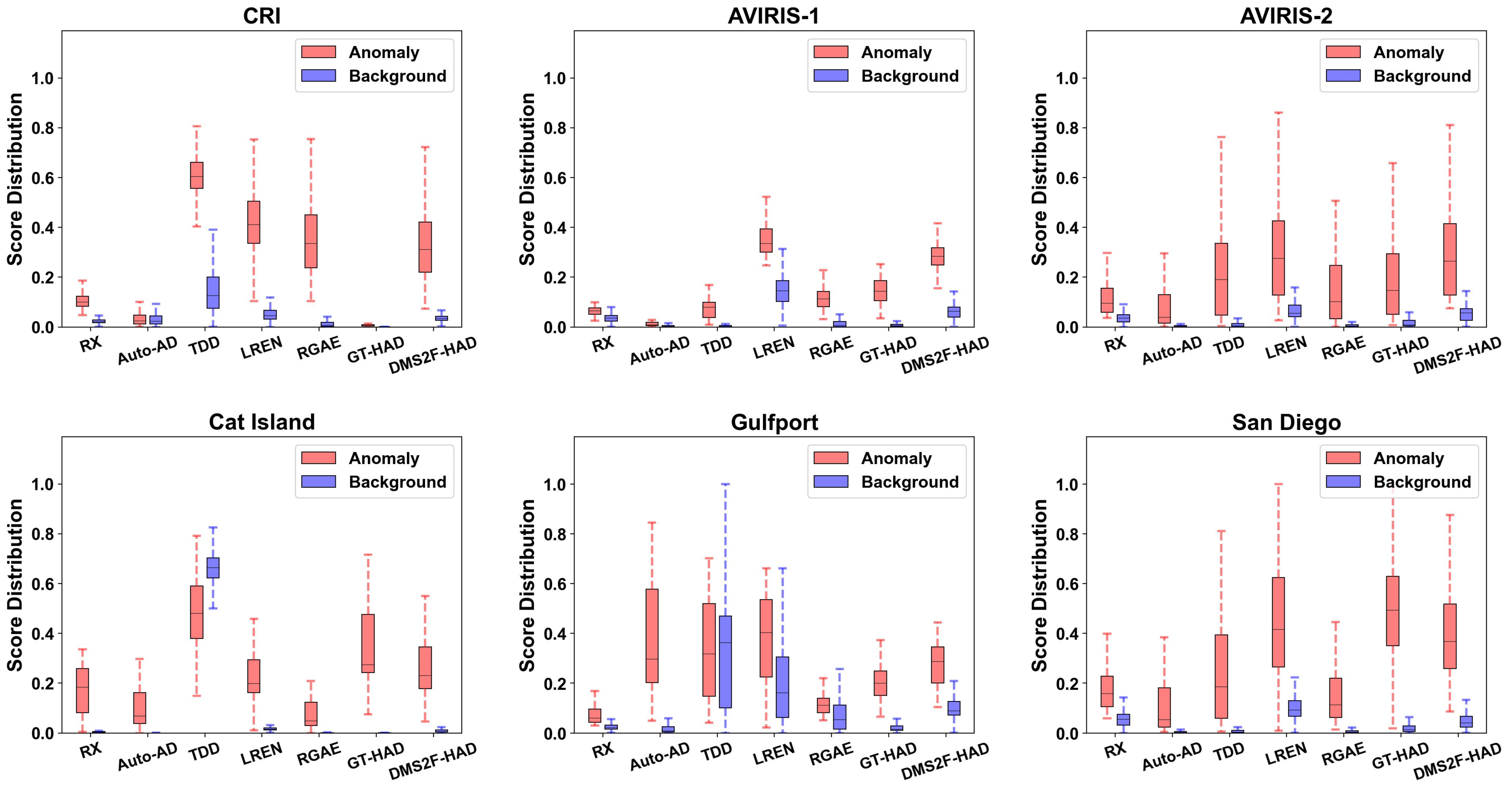}
    \caption{Box-whisker plots of anomaly scores of different HAD methods on six datasets.}
    \label{fig:box}
\end{figure*}

  In Figure~\ref{fig:anomaly}, we plot and compare the color anomaly maps of DMS2F-HAD along with the baseline methods. Our approach demonstrates superior detection capabilities, producing results that closely align with ground truth data. It effectively suppresses complex background textures in challenging environments such as CRI, San Diego, and AVIRIS-1. This visual clarity is further quantified via box-whisker plots as shown in Figure~\ref{fig:box}. Our model consistently demonstrates a clear separation with minimal overlap and exhibits tight background clustering. Although methods such as GT-HAD perform effectively on certain datasets, including San Diego and Cat Island, they frequently lack this degree of consistent background suppression, which is crucial for reliable detection. In contrast, traditional methods like RX and other deep models such as Auto-AD and TDD exhibit significant overlap, indicating an increased risk of false alarms. This performance is a direct result of our architectural design, where the dual-branch structure learns to model and suppress the background by processing spatial and spectral information independently, while the gated fusion mechanism integrates these features for a more precise final detection.

\begin{table}[hb!]
\centering
%

\small
\setlength{\tabcolsep}{2pt}
\resizebox{\columnwidth}{!}{%
\begin{tabular}{
 @{\extracolsep{\fill}}
  l
  S[table-format=1.4]
  S[table-format=1.4]
  S[table-format=1.4]
  S[table-format=1.4]
  S[table-format=1.4, detect-weight] 
  S[table-format=1.4, detect-weight]
  S[table-format=1.4, detect-weight]
   @{} %
}
\toprule
\textbf{Dataset} & {\textbf{RX}} & {\textbf{AUTO-AD}} & {\textbf{TDD}} & {\textbf{LREN}} & {\textbf{RGAE}} & {\textbf{GTHAD}} & {\textbf{Ours}} \\
\midrule
CRI& 0.9989 & 0.5123 & 0.9906 & 0.9976 & 0.9983 & 0.9952 & \textbf{0.9999} \\
Salinas& 0.8073 & 0.9908 & 0.2993 & 0.7665 & 0.9291 & 0.9474 & \textbf{0.9994} \\
San Diego& 0.9403 & 0.9471 & 0.9447 & 0.9704 & \textbf{0.9920} & 0.9815 & 0.9722 \\
Bay Champ & \textbf{0.9999} & 0.9990 & 0.2809 & 0.2387 & 0.8662 & 0.9997 & 0.9997 \\
Abu-urban-2& 0.9946 & 0.9852 & 0.1825 & 0.9823 & 0.9994 & 0.9944 & \textbf{0.9994} \\
Abu-urban-3& 0.9513 & \textbf{0.9734} & 0.9275 & 0.7992 & 0.8205 & 0.9703 & 0.9611 \\
Abu-urban-4& 0.9887 & 0.9901 & 0.8579 & 0.9252 & 0.9948 & 0.9887 & \textbf{0.9976} \\
Abu-urban-5& 0.9692 & 0.8459 & 0.8448 & 0.4369 & 0.9569 & 0.9443 & \textbf{0.9700} \\
AVIRIS-1& 0.8866 & 0.7954 & 0.9520 & 0.9757 & 0.9886 & 0.9862 & \textbf{0.9891} \\
AVIRIS-2& 0.9181 & 0.9034 & 0.9091 & 0.9342 & 0.9392 & 0.8970 & \textbf{0.9851} \\
Cat Island& 0.9807 & 0.9825 & 0.2016 & 0.9310 & 0.9393 & 0.9960 & \textbf{0.9998} \\
Gulfport& 0.9526 & 0.9798 & 0.5304 & 0.7774 & 0.7461 & 0.9869 & \textbf{0.9897} \\
Pavia& 0.9538 & 0.9695 & 0.2344 & 0.8977 & 0.9042 & \textbf{0.9994} & 0.9772 \\
Texas Coast& 0.9907 & 0.9897 & 0.1040 & 0.9834 & 0.9821 & \textbf{0.9971} & 0.9888 \\
\midrule
\rowcolor{gray!20}\textbf{Average}  & 0.9523 & 0.9189 & 0.5900 & 0.8297 & 0.9326 & 0.9774 & \textbf{0.9878} \\
\bottomrule
\end{tabular}

}
\caption{Anomaly detection results (AUC). The best performance per dataset is shown in \textbf{bold}.}
\label{tab:auc}
\end{table}

\subsubsection{Quantitative Analysis}

\begin{figure*}[!ht]
    \centering
    \includegraphics[width=0.8\textwidth]{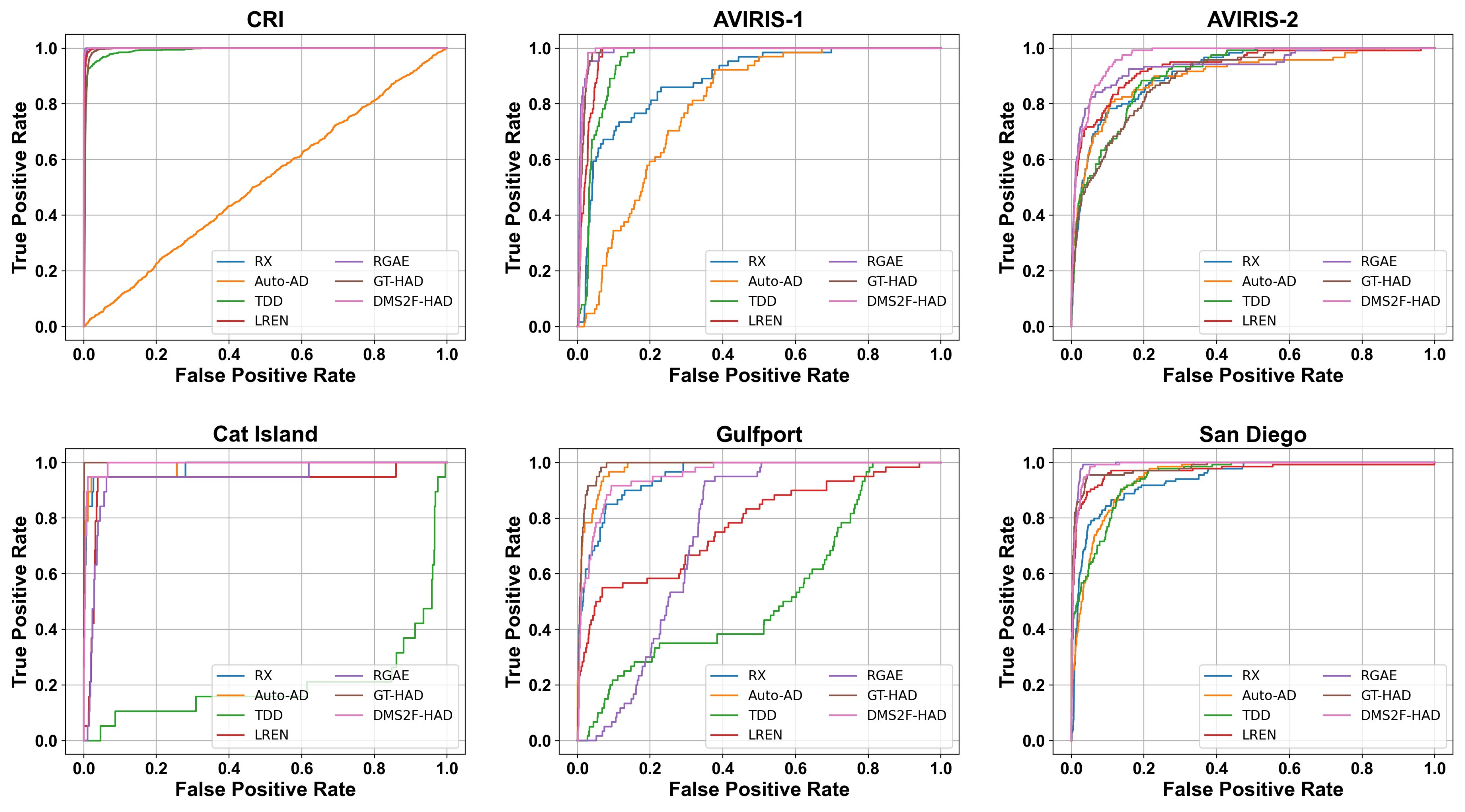} 
    \caption{2D ROC curves of seven different methods on six datasets.}
    \label{fig:roc}
\end{figure*}

The quantitative findings, detailed in Table~\ref{tab:auc}, clearly demonstrate the superiority of our model. DMS2F-HAD achieves an AUC of 0.9878, markedly outperforming the next leading method, GT-HAD, which records an AUC of 0.9774. Our model achieves the highest performance across 9 of the 14 datasets. The ROC curves illustrated in Figure~\ref{fig:roc} further corroborate this, showing that DMS2F-HAD consistently operates in the optimal top-left quadrant, indicative of a high detection rate with a minimal false alarm rate. This impressive performance underscores the effectiveness of our dual-branch architecture in capturing both spectral signatures and discriminative spatial contexts.

\subsubsection{Inference Time}
\begin{table}[ht!]

\small
\setlength{\tabcolsep}{4pt}
\begin{tabular}{
  l
  S[table-format=2.2]
  S[table-format=2.2]
  S[table-format=3.2]
  S[table-format=1.3, detect-weight] 
}
\toprule
\textbf{Dataset} & {\makecell{\textbf{AUTO-AD}}} & {\makecell{\textbf{TDD} }} & {\makecell{\textbf{GTHAD}}} & {\makecell{\textbf{DMS2F-HAD } \\ \textbf{(Ours)}}} \\
\midrule
CRI            & 29.23 & 11.70 & 290.13 & \textbf{2.335} \\
Salinas        & 7.54  & 1.57  & 19.57  & \textbf{0.405} \\
San Diego      & 8.66  & 1.52  & 10.20  & \textbf{0.316} \\
Bay Champagne  & 6.43  & 1.12  & 13.91  & \textbf{0.337} \\
Abu-urban-2    & 3.33  & 1.11  & 14.21  & \textbf{0.327} \\
Abu-urban-3    & 7.36  & 3.55  & 14.42  & \textbf{0.352} \\
Abu-urban-4    & 3.74  & 3.63  & 14.30  & \textbf{0.318} \\
Abu-urban-5    & 9.54  & 2.76  & 14.49  & \textbf{0.323} \\
AVIRIS-1      & 8.71  & 2.77  & 10.54  & \textbf{0.891} \\
AVIRIS-2      & 9.27  & 1.39  & 21.26  & \textbf{0.433} \\
Cat Island     & 8.63  & 1.40  & 33.84  & \textbf{0.583} \\
Gulfport       & 2.01  & 1.33  & 9.92   & \textbf{0.296} \\
Pavia          & 22.68 & 0.63  & 30.01  & \textbf{0.511} \\
Texas Coast    & 3.38  & 1.24  & 10.87  & \textbf{0.340} \\
\midrule
\rowcolor{gray!20}\textbf{Average} & 9.32 & 2.55 & 36.26 & \bfseries 0.555 \\
\bottomrule
\end{tabular}
\caption{Inference time (seconds). Only the methods with comparable implementations are considered.}
 \vspace{-6pt}
\label{tab:time}
\end{table} 
For practical HAD, computational efficiency is as crucial as the detection success rate. In this regard, we compare the inference time of DMS2F-HAD with other DL methods such as AUTO-AD, TDD, and GTHAD. From Table~\ref{tab:time}, we can see that DMS2F-HAD is able to beat all methods by a significant margin, with an average inference time of \textbf{0.55} seconds across 14 datasets. This is approximately $4.6\times$ faster than the next fastest model, TDD and over $65\times$ faster than GT-HAD. This remarkable efficiency stems from our lightweight Mamba-based design, which avoids the quadratic complexity of traditional transformers.

\subsubsection{Comparison with Mamba-based 
HAD} \label{sec:Mambaeval}

\begin{table}[ht!]

\small
\setlength{\tabcolsep}{2pt}
\begin{tabular*}{\columnwidth}{
  l
  @{\extracolsep{\fill}}
  S[table-format=1.4, detect-weight]
  S[table-format=1.4, detect-weight]
}
\toprule
\textbf{Dataset} & 
{\makecell[c]{\textbf{MMR-HAD (Reported)} \\ \small \textbf{Params}: 2.12M \\ \small \textbf{FLOPs}: 3.5G}} & 
{\makecell[c]{\textbf{DMS2F-HAD (Ours)} \\ \small \textbf{Params}: \textbf{0.64M} \\ \small \textbf{FLOPs}: \textbf{0.12G}}} \\
\midrule
Cat Island      & 0.9983 & \textbf{0.9998} \\
Bay Champagne   & 0.9972 & \textbf{0.9997} \\
Texas Coast     & \textbf{0.9958} & 0.9888 \\
Gulfport        & \textbf{0.9964} & 0.9897 \\
\bottomrule
\end{tabular*}
\caption{Comparison of AUC with SOTA model MMR-HAD \cite{10884568} on common datasets, including model complexity.}
\label{tab:auc_overlap}
\end{table}
We further benchmark DMS2F-HAD against the SOTA Mamba-based model, MMR-HAD \cite{10884568}, using four datasets common to both studies, as presented in Table \ref{tab:auc_overlap}. The results show that DMS2F-HAD achieves comparable state-of-the-art accuracy, with each method outperforming on two datasets. Importantly, our model delivers this level of accuracy with substantially greater efficiency, requiring $3.3\times$ fewer parameters and $29\times$ fewer FLOPs than MMR-HAD. This highlights efficiency as the key differentiator and establishes a new benchmark in the accuracy–efficiency trade-off, positioning DMS2F-HAD as a more practical solution for real-time and resource-constrained HAD applications.The substantial speedup is directly improving the theoretical advantages of our design. By replacing multi-head self-attention and complex hybrid blocks used in prior works with streamlined, linear-complexity Mamba blocks, we eliminate the primary computational bottlenecks.

\subsubsection{Ablation Study}

To validate our design, we analyze the contributions of the spatial and spectral branches and the gated fusion mechanism (Table \ref{tab:ablation_all}). The ``Spatial Only" model  achieves a strong  average AUC of 0.9776, significantly outperforming ``Spectral Only" model (0.9521). Notably, naive ``Addition Fusion" degrades performance to 0.9764 (0.12\% lower than the spatial baseline), suggesting that unweighted summation allows weaker features to corrupt discriminative ones. In contrast, our Gated Fusion achieves an average AUC of 0.9878. While the average gain is $\approx 1.1\%$, the impact is decisive in complex scenes. For instance, on the Gulfport dataset, Gated Fusion improves AUC by over 9\% compared to Addition Fusion (0.9897 vs 0.8979). This confirms that the adaptive gate is essential for generalizing across diverse environments where static summation fails.


\begin{table}[!ht]
\centering

 \small
\setlength{\tabcolsep}{3.5pt}
\begin{tabular}{
  l |
  S[table-format=1.4]
  S[table-format=1.4] |
  S[table-format=1.4]
  S[table-format=1.4, table-column-width=1.4cm, detect-weight] 
}
\toprule
\textbf{Dataset} & {\begin{tabular}{@{}c@{}}\textbf{Spectral} \\ \textbf{Only}\end{tabular}} & 
{\begin{tabular}{@{}c@{}}\textbf{Spatial} \\ \textbf{Only}\end{tabular}} & 
{\begin{tabular}{@{}c@{}}\textbf{Addition} \\ \textbf{Fusion}\end{tabular}} & 
{\begin{tabular}{@{}c@{}}\textbf{Gated Fus-} \\ \textbf{ion (Main)}\end{tabular}} \\
\midrule
CRI         & 0.9997 & 0.9990 & 0.9999 & \textbf{0.9999} \\
Salinas     & 0.9849 & 0.9970 & 0.9972 & \textbf{0.9994} \\
San Diego   & 0.9158 & 0.9674 & 0.9532 & \textbf{0.9722} \\
Bay Champagne& 0.9784 & 0.9991 & 0.9964 & \textbf{0.9997} \\
Abu-urban-2 & 0.9994 & 0.9993 & 0.9993 & \textbf{0.9994} \\
Abu-urban-3 & 0.9558 & 0.9272 & 0.9395 & \textbf{0.9611} \\
Abu-urban-4 & 0.9966 & 0.9960 & 0.9958 & \textbf{0.9976} \\
Abu-urban-5 & 0.9692 & 0.9692 & 0.9658 & \textbf{0.9700} \\
AVIRIS-1    & 0.9856 & 0.9716 & 0.9710 & \textbf{0.9891} \\
AVIRIS-2    & 0.8701 & 0.9755 & 0.9999 & \textbf{0.9851} \\
Cat Island  & 0.9704 & 0.9988 & 0.9981 & \textbf{0.9998} \\
Gulfport    & 0.7655 & 0.9410 & 0.8979 & \textbf{0.9897} \\
Pavia       & 0.9525 & 0.9584 & 0.9675 & \textbf{0.9772} \\
Texas Coast & 0.9859 & 0.9872 & 0.9882 & \textbf{0.9888} \\
\midrule
\rowcolor{gray!20}\textbf{Average} & 0.9521 & 0.9776 & 0.9764 & \textbf{0.9878} \\
\rowcolor{gray!20}\textbf{$\Delta$ vs. Spatial} & {-2.55\%} & \multicolumn{1}{c|}{(Baseline)}  & {-0.12\%} & \textbf{+1.02\%} \\

\bottomrule
\end{tabular}
\caption{Ablation study on the effectiveness of our dual-branch design and fusion mechanism. The gain ($\Delta$) is measured against the strongest single-branch baseline (Spatial Only).}
 \vspace{-12pt}
\label{tab:ablation_all}
\end{table}



\section{Conclusions}
\label{sec:remarks}


In this paper, we introduced \textbf{DMS2F-HAD}, a novel dual-branch Mamba-based network that efficiently models both long-range spatial and sequential spectral dependencies for hyperspectral anomaly detection. Our model establishes a new benchmark in the accuracy-efficiency trade-off, achieving a state-of-the-art average AUC of 98.78\%.By utilizing linear-complexity Mamba blocks without heavy attention mechanisms, our design is remarkably compact, requiring $3.3\times$ fewer parameters than the competing SOTA Mamba model while delivering an average inference time of 0.55 seconds operating at $4.5\times$ faster than the next leading method. These results position DMS2F-HAD as a significant step toward real-time practical HAD. Moreover, the adaptive gated fusion mechanism ensures consistent generalization across diverse terrains, and complex background clutter where traditional fusion methods fail. 



\section*{Acknowledgment}
\label{sec:ack}
This work is supported by the Air Force Office of Scientific Research under award number FA2386-23-1-4003 and Deakin University.


{
    \small
    \bibliographystyle{ieeenat_fullname}
    \bibliography{main}
}

\end{document}